# Benchmark Analysis of Various Pre-trained Deep Learning Models on ASSIRA Cats and Dogs Dataset

**Galib Muhammad Shahriar Himel[1,2,4], Md. Masudul Islam[3,4]**
[1]Universiti Sains Malaysia, 11800 USM Penang, Malaysia
[2]American International University-Bangladesh, Dhaka-1229, Bangladesh
[3] Bangladesh University of Business and Technology, Dhaka-1216, Bangladesh
[4]Jahangirnagar University, Dhaka-1342, Bangladesh

**Corresponding author:** Galib Muhammad Shahriar Himel (galib.muhammad.shahriar@gmail.com).

**ABSTRACT** As the most basic application and implementation of deep learning, image classification has grown in popularity. Various datasets are provided by renowned data science communities for benchmarking machine learning algorithms and pre-trained models. The ASSIRA Cats & Dogs dataset is one of them and is being used in this research for its overall acceptance and benchmark standards. A comparison of various pre-trained models is demonstrated by using different types of optimizers and loss functions. Hyper-parameters are changed to gain the best result from a model. By applying this approach, we have got higher accuracy without major changes in the training model. To run the experiment, we used three different computer architectures: a laptop equipped with NVIDIA GeForce GTX 1070, a laptop equipped with NVIDIA GeForce RTX 3080Ti, and a desktop equipped with NVIDIA GeForce RTX 3090. The acquired results demonstrate supremacy in terms of accuracy over the previously done experiments on this dataset. From this experiment, the highest accuracy which is 99.65% is gained using the NASNet Large.

## I. INTRODUCTION

Deep Learning models have become very popular for image processing. The benchmark analysis is extremely important to verify whether the performance of a model is good or not. Various such researches [1][2][3][4][5] have been published in recent years. Each of the datasets used in such research follows a benchmark standard. Various datasets are used to measure the performance of these models, but not all datasets conform to benchmark standards. Among those that do meet the benchmark standards, the datasets of ASIRRA Cats & Dogs [6] are particularly noteworthy. In the past few years, several experiments have been conducted on the cat and dog datasets, and some notable achievements have been made. Brownlee [7] achieved 97% accuracy using the fine-tuning technique of the VGG16 model, Manish [8] achieved 96.67% accuracy using VGG16 Fine Tuning, and William [9] achieved 99.1% using the Ensemble ResNet method, where the ensemble ResNet18 and ResNet34 models. However, it should be noted that not all datasets are benchmarked according to the same standard. Observing the experiments, it is clear that using pre-trained models, the training can be done more accurately rather than using a custom model in which the model is only trained by the specific dataset. However, the pre-trained models have already been trained on a huge dataset; in most cases trained on the ImageNet [10] dataset which contains 14,197,122 images. That is why the neurons in the convolutional neural network are already pre-assigned with trained weights. This makes further training more efficient. Another observation is that there was no significant manipulation of loss functions and optimizers. The objective of this work is to perform a comprehensive comparison among various machine learning models by detailed manipulation of the loss function and optimizer, to achieve maximum accuracy. Additionally, we also evaluate the run-time performance of different models on various GPU architectures. We can consider our results as state-of-the-art and analyze them accordingly. Further, some important findings are: 1) The correlation between model complexity and accuracy is not linear. 2) Model accuracy slightly depends on input image size. The more the dimension of the input image is, the more the accuracy will probably be. As it is easier for a model to extract efficient features from larger images. 3) After reaching a threshold, there is no further increase in recognition accuracy. 4)

Variation of GPU architecture doesn't have any effect on the accuracy. However, the Run-time for different models varies in different GPUs because of their memory management architecture. 5) In most cases Binary Cross Entropy and Hinge loss functions provide the highest accuracy with some exceptions. The rest of the paper is structured in the following manner: Section II outlines the relevant research; Section III provides a brief overview of the implementation architectures considered; Section IV describes the measured results using graphical representation; Section V reports and discusses the achieved results; and Section VI presents our conclusion.

## II. RELATED WORKS

There is some significant researches experimental work in image classifications especially on ASIRRA Cats & Dogs dataset. Kseniya [11] achieved 70.8% accuracy using the fuzzy net on VGG finetuning model. In 2018 [12], research regarding Cats & Dogs classification was done to improve the CNN performance level and gained 88.31% of accuracy. In 2019, Enzhi, and Chunyang [13] used the LeNet-5 CNN model on Cats vs. Dogs, Cifar-10, and Fer2013 datasets and achieved 91.89% accuracy. They improved the result from previously done experiments using the Trans-convolution Neural Network algorithm [14] on a similar dataset. Tejasv & Himanshu [15] made a comparison among VGG16, MobileNet, Resnet50, and InceptionV3 on the Cats & Dogs dataset and describe various facts from the observation. In 2015, using the AlexNet [16] model in the ImageNet competition [17] the error rate was 15.3% whereas Tejasv & Himanshu were able to reduce the error rate to 2.5%. In 2020, Yao Yu [18] developed a customized deep neural network to achieve an accuracy of 92.7% on the Cats & Dogs dataset. They designed a model mainly composed of stacked residual layers and a spatial transformer network (STN) [19]. In 2020, Deshan & Christos [20] have shown the impact of data augmentation, feature extraction, and fine-tuning on the accuracy of the model. The accomplished results show that having a small dataset, those approaches are very effective when dealing with image data. Youngjun [21] in 2021, implemented SVM and CNN with data augmentation and without augmentation while using different activation functions to show a comparison among the results. The best result was gained using the LeakyReLU activation function.

## III. IMPLEMENTATION AND METHOD

### A. DATASET

We used the ASIRRA Cats & Dogs dataset which contains 25,000 images. Asides from the garbage data the total number of usable image data is 24,994. In **Table 1**, we demonstrate the data split. The dataset is divided into three portions: train images (12492), validation images (6251), and test images (6251). These images exhibit considerable diversity in terms of quality and size.

TABLE 1.
DATASET DESCRIPTION

| Total images | Train images | Val images | Test images |
|---|---|---|---|
| Cats | 6246 | 3125 | 3126 |
| Dogs | 6246 | 3126 | 3125 |
| Grand Total | 12492 | 6251 | 6251 |

### B. BENCHMARKING

We have implemented the standardized method for Deep Neural Networks in Python. The Keras [22] and TensorFlow [23] packages are used for neural network processing with cuDNN-v8.3.2 [24] and CUDA-v11.4 [25] as the back end. We run all the experiments on two laptops and a desktop system:

1) The desktop is equipped with an Intel Core i7-12700K CPU @ 3.61 GHZ, 32GB DDR4 RAM 5600 MHz, and NVIDIA GeForce RTX 3090. The operating system is Windows 11.
2) The laptop is equipped with an Intel Core i9-12900H CPU @ 2.90 GHZ, 64GB DDR4 RAM 4800 MHz, and NVIDIA GeForce RTX 3080Ti. The operating system is Windows 11.
3) The laptop is equipped with an Intel Core i7-7700HQ CPU @ 2.80 GHZ, 32GB DDR4 RAM 2400 MHz, and NVIDIA GeForce GTX 1070. The operating system is Windows 10.

### C. PERFORMANCE INDICES

To ensure a fair and direct comparison, we replicate the same sampling policies by obtaining models trained with TensorFlow and also by converting models trained with other deep learning frameworks to TensorFlow. All pre-trained models [26] require input images to be normalized consistently. Specifically, they require mini-batches of RGB images with dimensions of 3 × H × W, where the values of H and W are:

- 331 pixels for the NASNet-Large model;
- 224 pixels for all the other models considered;

We believe that measuring multiple performance indicators is important for a thorough evaluation of DNN models. Different performance indicators are described in a detailed manner in the result section.

### D. SYSTEM ARCHITECTURE

This segment provides a brief overview of the pre-trained models. We have chosen several models, some with a focus on time effectiveness, while others prioritize efficiency & performance.

1) MODELS:

We consider the following models: Xception [27], VGG16 [28], VGG19 [29], ResNet50 [30], InceptionV3 [31], InceptionResNetV2 [32], MobileNet [33], MobileNetV2 [34], DenseNet121 [35], and NASNetLarge [36].

1) *Xception*: Xception is a deep convolutional neural network architecture. It aims to improve model efficiency and performance by replacing standard convolutional layers with depth-wise separable convolutions. It was trained on the ImageNet dataset, which consists of over a million labeled images across 1000 categories. The model size is 88 MB, the number of parameters is 22.9M and the depth is 81.

2) *VGG16*: VGG16 is a popular convolutional neural network architecture that has been pre-trained on a large dataset of images called ImageNet. It consists of 16 layers, with a series of convolutional layers followed by fully connected layers. Its performance has been proven to be very good, with high accuracy on many benchmark datasets. The model size is 528 MB, the number of parameters is 138.4M and the depth is 16.
3) *VGG19*: VGG19 is a convolutional neural network architecture that has 19 layers, including 16 convolutional layers and 3 fully connected layers. It is a popular deep-learning model for image classification tasks due to its exceptional performance on the ImageNet dataset. VGG19 has over 143.7 million parameters with a size of 549 MB and is trained on a large dataset of over 1 million images. The model has a simple architecture, with small convolution filters (3x3) and max pooling layers, which helps to achieve good accuracy on various computer vision tasks.
4) *ResNet50*: ResNet50 is a deep convolutional neural network architecture that has 50 layers and was first introduced in 2015. It has been pre-trained on a large dataset (ImageNet) to perform image recognition tasks. ResNet50 is known for its residual blocks, which allow for the network to be trained more effectively and efficiently, ultimately leading to better accuracy. The model size is 98 MB, the parameter number is 25.6M and the depth is 107.
5) *InceptionV3*: InceptionV3 is a convolutional neural network (CNN) architecture that has been pre-trained on the ImageNet dataset for image classification tasks. It uses a combination of 1x1, 3x3, and 5x5 convolutions with max pooling layers to create a deep and complex network that can efficiently process images while minimizing the number of parameters. The model size is 92 MB, the parameter number is 23.9M and the depth is 189.
6) *InceptionResNetV2*: InceptionResNetV2 is a deep convolutional neural network architecture that is used for image classification tasks. It is a combination of the Inception and ResNet architectures, and it has been trained on a large dataset of images, such as the ImageNet dataset, to learn a rich set of features that can be used for various computer vision tasks. The model has achieved state-of-the-art results in several image classification challenges and is widely used as a starting point for transfer learning in various applications. The model size is 215 MB, the parameter number is 55.9M and the depth is 449.
7) *MobileNet*: The model size is 16 MB; the parameter number is 4.3M and the depth is 55. MobileNet is a deep neural network architecture designed for efficient processing on mobile and embedded devices. It uses depth-wise separable convolution, which reduces computational complexity while maintaining accuracy, making it well-suited for real-time applications with limited computing resources.
8) *MobileNetV2*: MobileNetV2 is a neural network architecture designed for mobile and embedded vision applications. It is based on depth-wise separable convolutions, which enable efficient model inference on resource-constrained devices. The model size is 14 MB, the parameter number is 3.5M and the depth is 105.
9) *DenseNet121*: DenseNet121 is a convolutional neural network (CNN) that has been pre-trained on a large dataset of images. It uses a unique architecture where each layer is connected to every other layer in a dense block, promoting feature reuse and reducing the number of parameters. The model size is 33 MB, the parameter number is 8.1M and the depth is 242.
10) *NASNetLarge*: NASNetLarge is a deep convolutional neural network architecture developed through neural architecture search (NAS) by Google AI researchers. It consists of 88.9 million parameters and is pre-trained on the ImageNet dataset for image classification tasks. NASNetLarge has achieved state-of-the-art performance on multiple benchmarks and has been shown to generalize well to various computer vision tasks. It is known for its high accuracy and efficiency in terms of computational resources compared to other CNN architectures. The model size is 343 MB and the depth is 533.

The general architecture of our experiment is given in **Figure 1**.

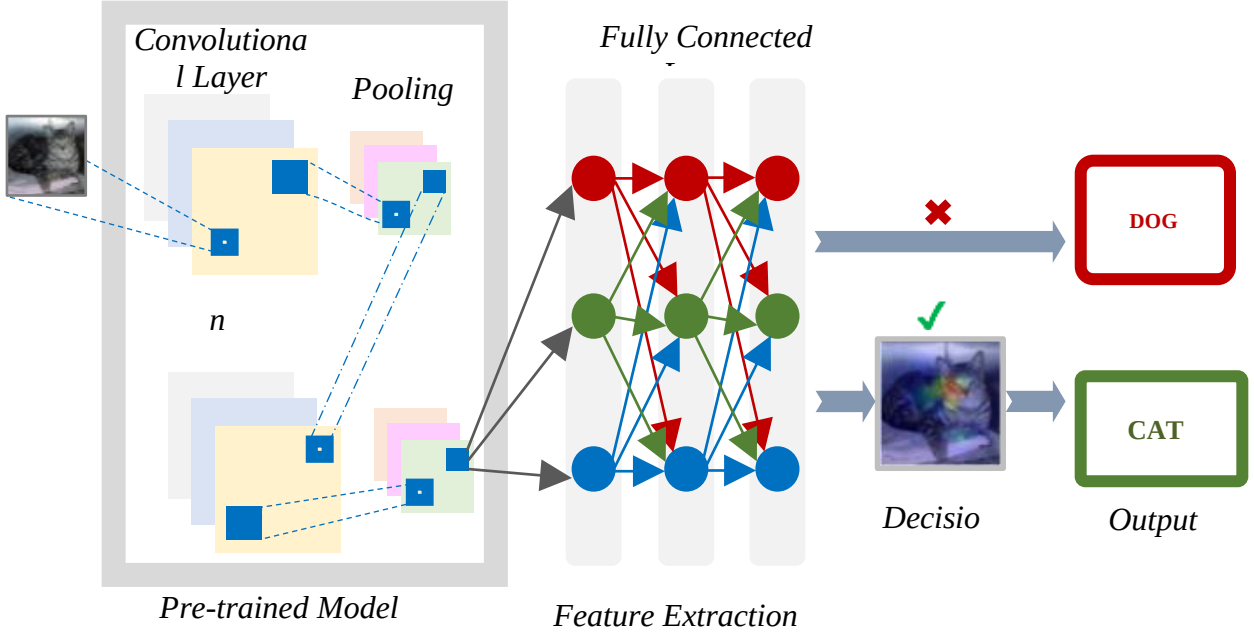

**FIGURE 1. General description of the training method. Here in the pre-trained model section, ten models are used sequentially**

### 2) OPTIMIZER:

We have used four optimizers for each model—Adam, Adamax, RMSprop, and SGD.

1) *Adam*: Adam (short for Adaptive Moment Estimation) is a gradient-based optimization algorithm used in deep learning for updating the weights of a neural network during training. It combines ideas from both Momentum and RMSprop optimization methods. The mathematical equation of the Adam optimizer is:

$$m_t = \beta_1 m_{t-1} + (1-\beta_1)\left[\frac{\delta L}{\delta w_t}\right] \vartheta_t = \beta_2 \vartheta_{t-1} + (1-\beta_2)\left[\frac{\delta L}{\delta w_t}\right]^2$$

2) *Adamax*: AdaMax is a generalization of Adam from the $l_2$ norm to the $l_\infty$ norm. Define:

$$u_t = \max\left(\beta_2 \cdot v_{t-1}, \left|\left[\frac{\delta L}{\delta w_t}\right] \vartheta_t\right|\right)$$

We can plug into the Adam update equation by replacing $\sqrt{\frac{v_t}{1-\beta_2^t}} + \epsilon$ with $u_t$ to obtain the AdaMax update rule:

$$\theta_{t+1} = \theta_t - \frac{\eta}{u_t} \cdot \frac{m_t}{1-\beta_1^t}$$

3) *RMSprop*: RMSProp tackles this by keeping a moving average of the squared gradient and adjusting the weight updates by this magnitude. The gradient updates are performed as:

$$E[g^2]_t = \gamma E[g^2]_{t-1} + (1-\gamma) g_t^2$$
$$\theta_{t+1} = \theta_t - \frac{\eta}{\sqrt{E[g^2]_t + \epsilon}} g_t$$

4) *SGD*: Stochastic Gradient Descent is an iterative optimization technique that uses mini-batches of data to form an expectation of the gradient, rather than the full gradient using all available data. That is for weights ***w*** and a loss function ***L*** we have:

$$w_{t+1} = w_t - \eta \widehat{\nabla_w} L(w_t)$$

### 3) LOSS FUNCTIONS:

We have used four loss functions for every optimizer used in each model. These are—Binary Cross Entropy, Categorical Cross Entropy, Hinge, and Kullback Leibler Divergence. BCE and Hinge are used for Binary classification whereas CCE and KLD are generally used for multi-class classification. However, they can be also used for Binary classification.

*1. Binary Cross Entropy*: In binary classification, where the number of classes **M** equals 2, cross-entropy can be calculated as:

$$-(y \log(p) + (1-y)\log(1-p))$$

*2. Categorical Cross Entropy*: if there are more than two classes, we define a new term known as categorical cross-entropy. It is calculated as a sum of separate losses for each class label per observation. Mathematically, it is given as the following equation:

$$-\sum_{c=1}^{M} y_{o,c} \log(p_{o,c})$$

*3. Hinge*: Also known as Support Vector Machine (SVM) loss. The formula is:

$$\max(0, 1 - y \cdot \hat{y})$$

*4. Kullback Leibler Divergence*: Also known as Root Mean Square Error. The equation is as follows:

$$\text{RMSE} = \sqrt{\frac{1}{m}\sum_{i=1}^{m}(h(x^{(i)}) - y^{(i)})^2}$$

### E. EXPERIMENTAL SETUP

For our experiment, we used pre-trained models; all of which are trained on the ImageNet dataset which consists of a total of 1000 categories. That is why every pre-trained model architecture has an output layer of 1000 nodes. However, in our dataset, we only have two categories, and due to this we freeze the final layer & define a custom output layer of two nodes.

To effectively train the models with our dataset we have implemented data augmentation to increase the dataset. The data augmentation parameters, the hyperparameter settings, and the total combination of experiments are mentioned in **Table 2**, **Table 3**, and **Table 4**.

| Parameter | Value |
|---|---|
| Rescale | 1./255 |
| Zoom range | 0.3 |
| Rotation range | 15 |
| Horizontal flip | TRUE |

TABLE 2
DATA AUGMENTATION

TABLE 3
HYPERPARAMETER TUNING

| Hyperparameter | Optimization Space |
|---|---|
| Epoch | 25 |
| Batch size | 32 |
| Learning rate | 0.0001 |
| Optimizer | 'Adam', 'Adamax', 'RMSprop', 'SGD' |
| Loss function | [Binary cross entropy], [Categorical cross entropy], [Hinge], [Kullback Leibler Divergence] |
| Activation Function | 'softmax' |
| Class mode | 'categorical' |

TABLE 4
TOTAL COMBINATIONS OF EXPERIMENTS

| Pretrained Model | Optimizer | Loss function | Calculation | Total combination |
|---|---|---|---|---|
| 10 | 4 | 4 | 10×4× 4 | 160 |

## IV. RESULTS & DISCUSSION

### A. ACCURACY VS. LOSS

The training curve in following **Figure 2** represents the training and validation loss as well as the training and validation accuracy respectively. The detailed results are demonstrated in **Table 6**. 16 combinations of the experiment were run for each model; only the best one from each model is shown in **Figure 2**.

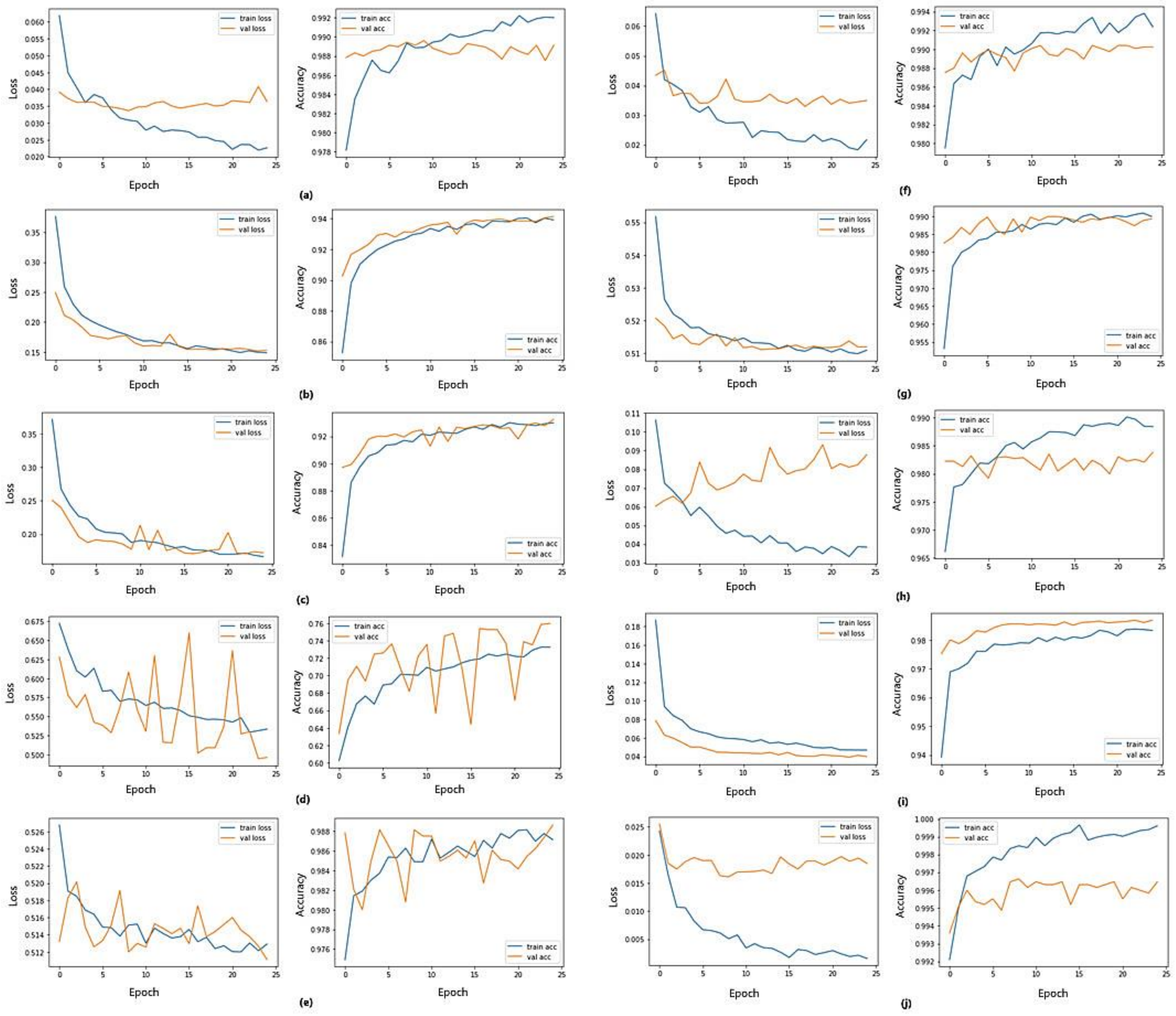

**FIGURE 2. Loss and Accuracy graph for (a) Xception, (b) VGG16, (c) VGG19, (d) ResNet50, (e) InceptionV3, (f) InceptionResNetV2, (g) MobileNet, (h) MobileNetV2, (i) DenseNet121, and (j) NASNetLarge**

### B. TEST ACCURACY COMPARISON OF MODELS

From every model, the combinations which provide the test accuracy are illustrated in **Table 5**.

TABLE 5
BEST TEST ACCURACY & ROC

| Model | Optimizer | Loss Function | Test Accuracy | ROC |
|---|---|---|---|---|
| *Xception* | Adamax | BCE | 0.9891 | 0.9987 |
| *VGG16* | Adam | BCE | 0.9414 | 0.9862 |
| *VGG19* | Adam | KLD | 0.9324 | 0.9828 |
| *ResNet50* | Adam | KLD | 0.7611 | 0.8453 |
| *InceptionV3* | Adamax | Hinge | 0.9886 | 0.9990 |
| *InceptionResNetV2* | Adamax | CCE | 0.9902 | 0.9988 |
| *MobileNet* | Adamax | Hinge | 0.9892 | 0.9989 |
| *MobileNetV2* | RMSprop | BCE | 0.9838 | 0.9971 |
| *DenseNet121* | SGD | BCE | 0.9869 | 0.9991 |
| *NASNetLarge* | Adamax | BCE | **0.9964** | **0.9996** |

**Figure 3**, shows the bar chart comparing the test accuracies of different models.

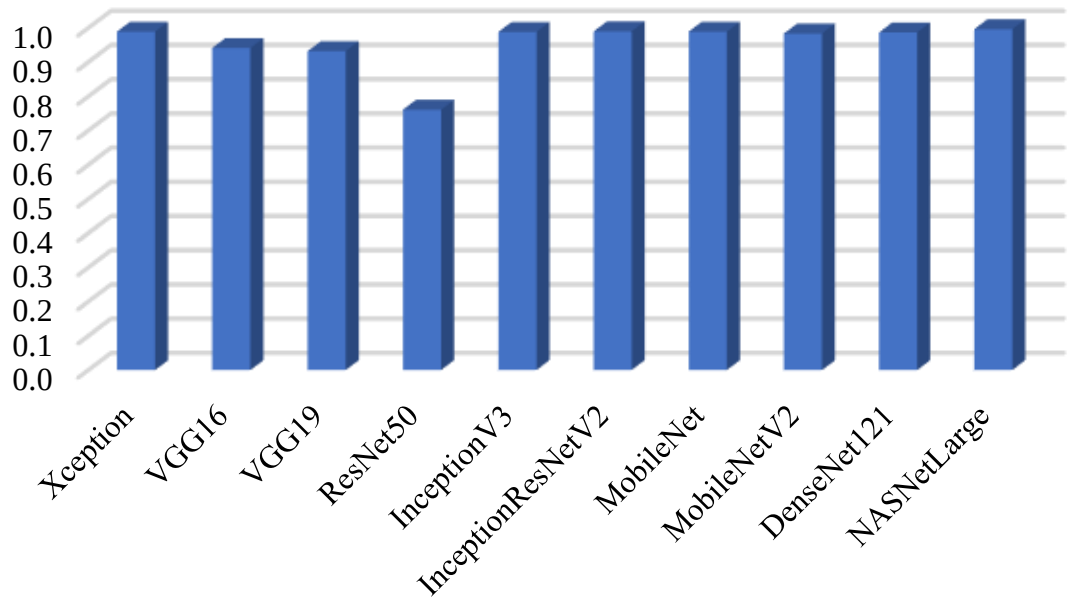

**FIGURE 3. Test Accuracy Comparison**

## C. MODEL ACCURACY VS. MODEL SIZE

**Figure 4** indicates a bubble chart of accuracy Vs. model size.

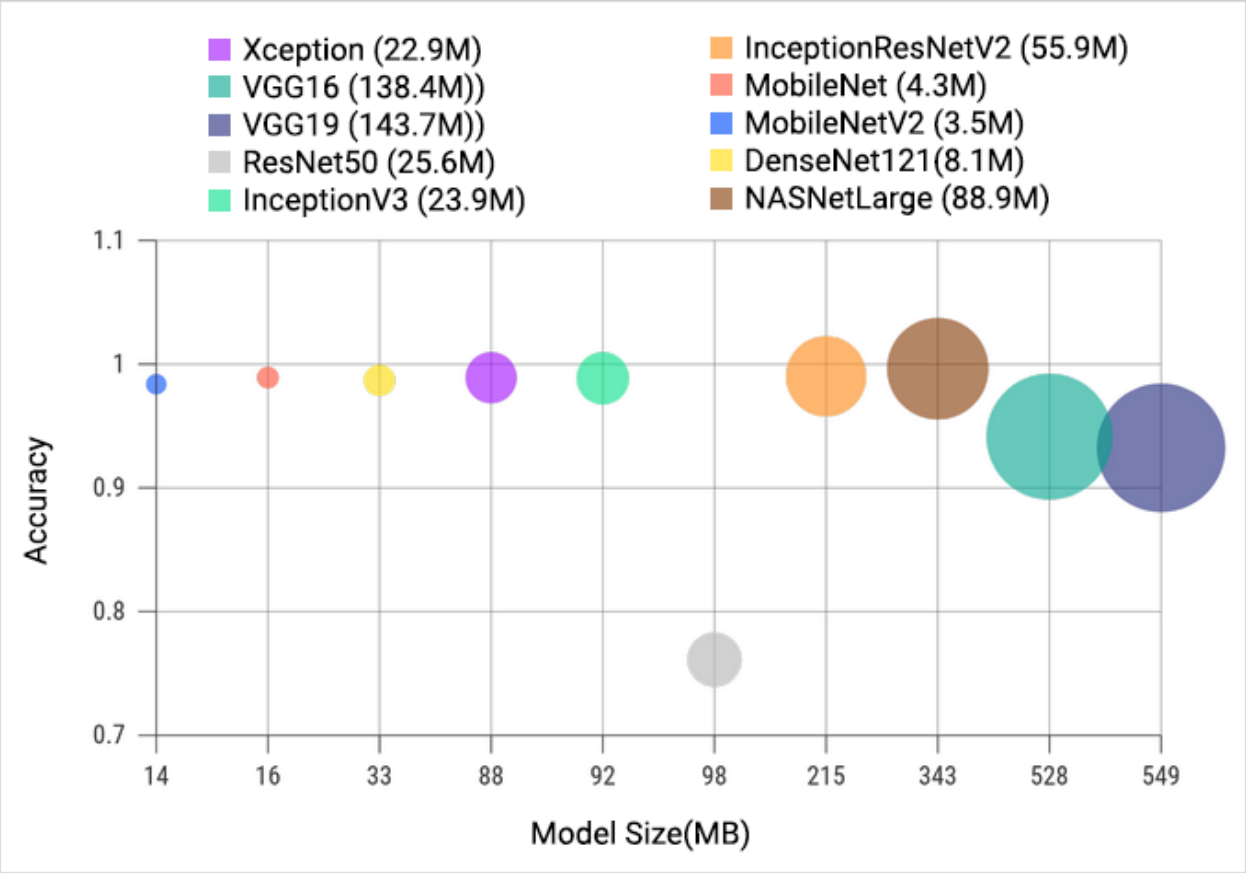

FIGURE 4. **Accuracy Vs. Model Size**

## D. ROC COMPARISON FOR ALL MODELS

**Figure 5** shows the ROC curve for all models applied in this experiment.

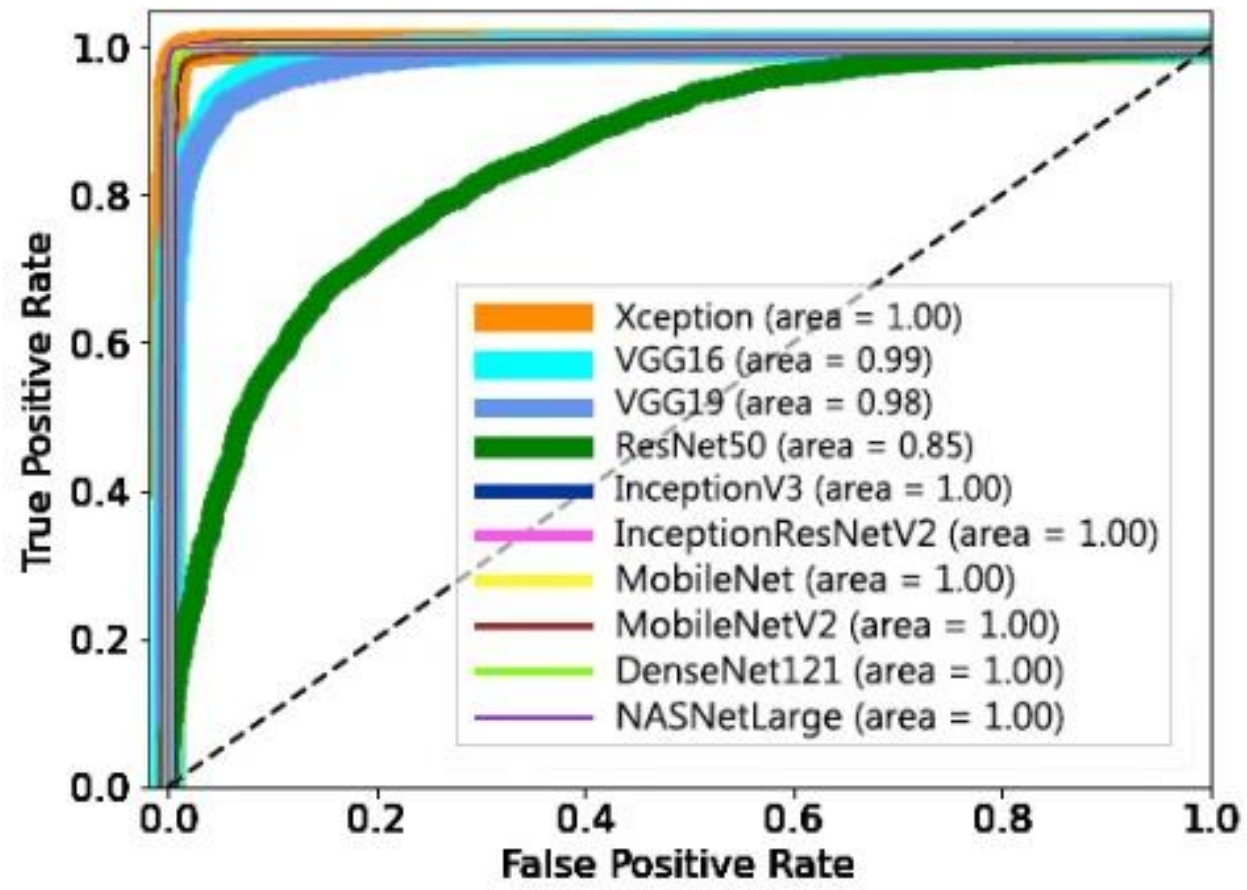

FIGURE 5. **ROC for all models**

## E. CLASSIFICATION REPORT

**Table 7** shows the precision, recall and F-1 scores of the best combination for every model. Precision is one indicator model's performance where it defines the quality of positive prediction rate of a model. From the table we can see that the true prediction rate for cats is slightly higher than that of dogs. As for recall, false positive rate for dogs is higher than that of cats as more cats are falsely identified as dogs.

## F. 5-FOLD CROSS-VALIDATION

We divided our dataset into five subsets or folds of approximately equal size. The model is trained on four of the folds and evaluated on the remaining fold, with the process repeated for all possible combinations of training and validating folds. **Figure 6** shows a Box-Whisker plot to make a comparison among all models. Among the ten models,

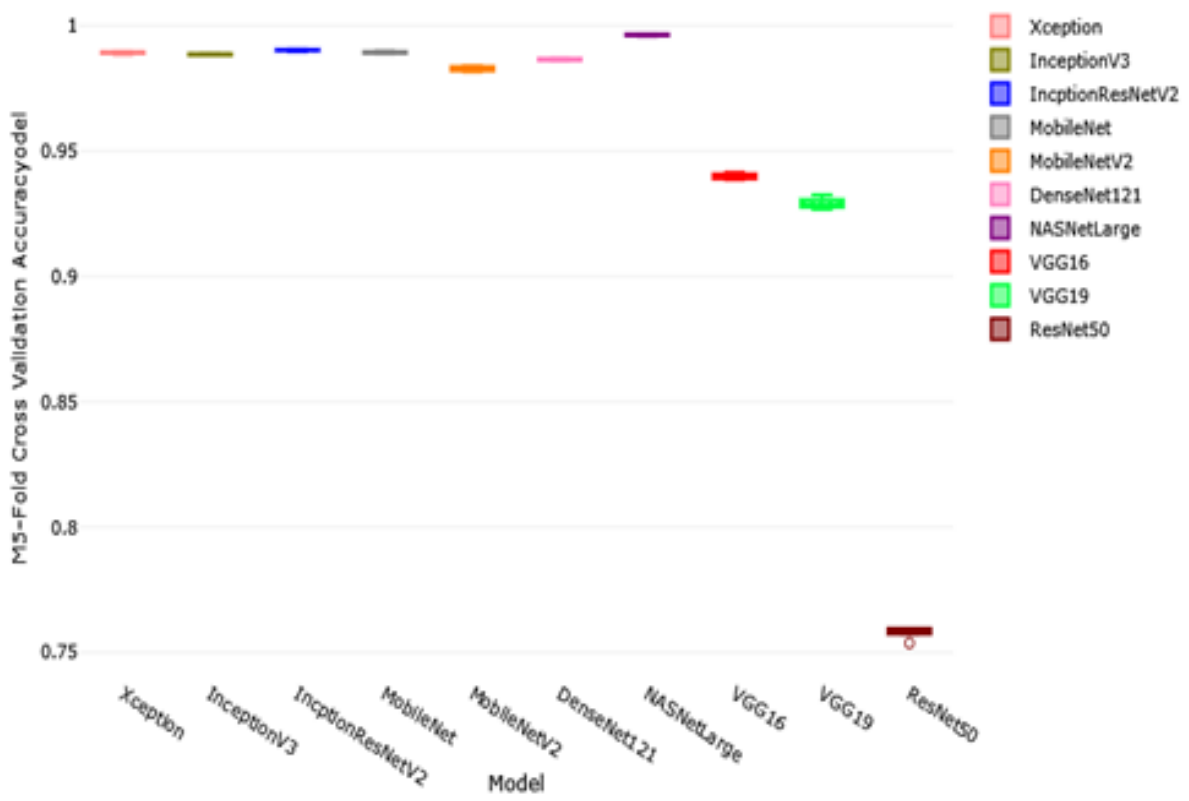

FIGURE 6. **Box-Whisker plot for 5-Fold Cross Validation**

NASNetLarge outperforms the other ones. **Figure 7** is illustrating the ROC curves of every fold for NASNetLarge.

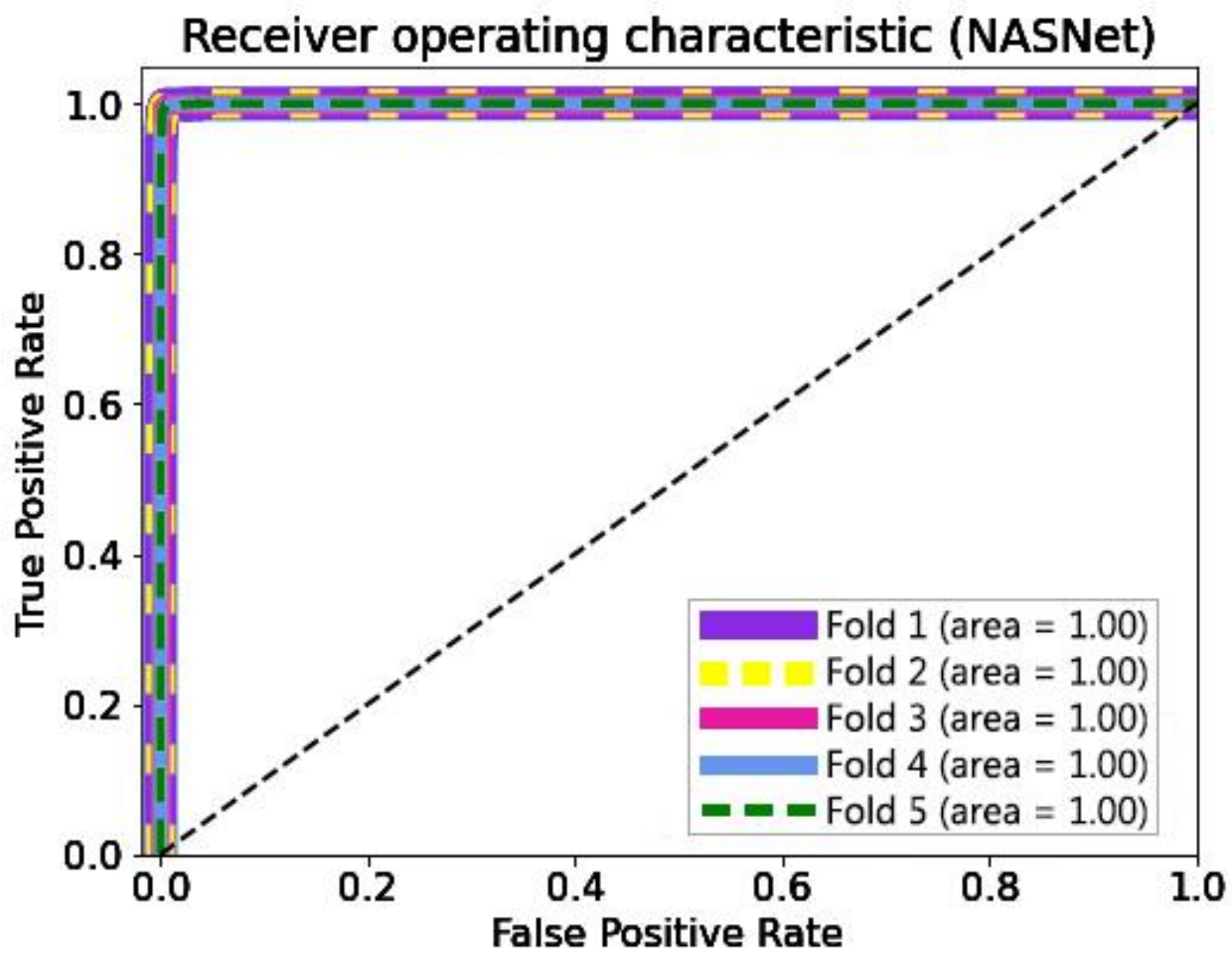

FIGURE 7. **5-Fold Cross Validation ROC for NASNet Large**

## G. CONFUSION MATRIX

**Figure 8** denotes confusion matrices of the best result from every set of combinations of pre-trained models. In the confusion matrices, true labels are denoted across y-axis and predicted labels are denoted across x-axis. The scale besides every confusion matrix denotes heatmap which indicates the sensitivity for the true and false positive rate.

## H. GradCAM EXPLANATION

**Figure 9**, describes the GradCAM [37] (Gradient-weighted Class Activation Mapping) experiment result where we observed that most of the features extracted to identify a specific class are almost similar in nature with some exceptions. However, these exceptions made a difference in accuracy. For example, the NASNetLarge model extracts features from eye-to-eye distance ratio, eye-to-ear distance ratio, eye to nose distance ratio more accurately than the other models; resulting higher prediction rate.

TABLE 6

DETAILED ACCURACY & LOSS SCORES OF EVERY POSSIBLE COMBINATION

| Loss Function ➡ | | BCE | | | | CCE | | | | Hinge | | | | KLD | | | |
|---|---|---|---|---|---|---|---|---|---|---|---|---|---|---|---|---|---|
| Model⬇ | Optimizer⬇ | Train Loss | Train Accuracy | Valid Loss | Valid Accuracy | Train Loss | Train Accuracy | Valid Loss | Valid Accuracy | Train Loss | Train Accuracy | Valid Loss | Valid Accuracy | Train Loss | Train Accuracy | Valid Loss | Valid Accuracy |
| Xception | Adam | 0.0246 | 0.9927 | 0.074 | 0.9848 | 0.0452 | 0.99 | 0.1119 | 0.9867 | 0.5121 | 0.9879 | 0.5124 | 0.9877 | 0.0433 | 0.9911 | 0.0886 | 0.9877 |
| | Adamax | 0.0226 | 0.992 | 0.0363 | **0.9891** | 0.021 | 0.9924 | 0.04 | 0.9878 | 0.5099 | 0.9909 | 0.5124 | 0.9888 | 0.0208 | 0.9927 | 0.0396 | 0.9885 |
| | RMSprop | 0.029 | 0.9911 | 0.0649 | 0.9874 | 0.0385 | 0.9909 | 0.1034 | 0.9856 | 0.5105 | 0.9898 | 0.5114 | 0.9886 | 0.0399 | 0.9909 | 0.0875 | 0.9875 |
| | SGD | 0.0456 | 0.9832 | 0.0384 | 0.9885 | 0.0365 | 0.9869 | 0.0358 | 0.9882 | 0.5208 | 0.9831 | 0.5164 | 0.9862 | 0.0354 | 0.9864 | 0.0362 | 0.9878 |
| VGG16 | Adam | 0.1489 | 0.9392 | 0.153 | **0.9414** | 0.1397 | 0.9427 | 0.1572 | 0.9398 | 0.5717 | 0.9382 | 0.5693 | 0.9365 | 0.1395 | 0.9426 | 0.156 | 0.9395 |
| | Adamax | 0.1834 | 0.9261 | 0.1675 | 0.9322 | 0.1658 | 0.9335 | 0.1587 | 0.9376 | 0.5897 | 0.9261 | 0.5799 | 0.9306 | 0.1655 | 0.9338 | 0.1599 | 0.936 |
| | RMSprop | 0.1509 | 0.9386 | 0.156 | 0.94 | 0.1475 | 0.9399 | 0.1762 | 0.9347 | 0.5717 | 0.9375 | 0.5693 | 0.936 | 0.1493 | 0.938 | 0.1623 | 0.9405 |
| | SGD | 0.343 | 0.87 | 0.3052 | 0.8871 | 0.2646 | 0.8932 | 0.23 | 0.908 | 0.6924 | 0.8644 | 0.6677 | 0.88 | 0.2643 | 0.894 | 0.2311 | 0.9095 |
| VGG19 | Adam | 0.177 | 0.9281 | 0.1719 | 0.9291 | 0.1672 | 0.9329 | 0.1812 | 0.9248 | 0.5876 | 0.9243 | 0.586 | 0.9227 | 0.1666 | 0.93 | 0.1721 | **0.9325** |
| | Adamax | 0.2144 | 0.9114 | 0.1898 | 0.9229 | 0.1908 | 0.9217 | 0.1791 | 0.9243 | 0.6074 | 0.9097 | 0.5946 | 0.9175 | 0.1942 | 0.9208 | 0.1797 | 0.9234 |
| | RMSprop | 0.1822 | 0.9255 | 0.177 | 0.9263 | 0.1752 | 0.9299 | 0.1781 | 0.9277 | 0.588 | 0.923 | 0.5821 | 0.9237 | 0.1821 | 0.9249 | 0.1884 | 0.9274 |
| | SGD | 0.3765 | 0.8499 | 0.3302 | 0.8723 | 0.2948 | 0.8786 | 0.2564 | 0.8947 | 0.7141 | 0.8392 | 0.6851 | 0.8632 | 0.2996 | 0.8755 | 0.2604 | 0.8914 |
| ResNet50 | Adam | 0.54 | 0.7296 | 0.5238 | 0.7399 | 0.5391 | 0.7301 | 0.5208 | 0.7501 | 0.819 | 0.6901 | 0.7925 | 0.7148 | 0.5335 | 0.7324 | 0.4962 | **0.7596** |
| | Adamax | 0.5638 | 0.7154 | 0.534 | 0.7341 | 0.5547 | 0.7211 | 0.5196 | 0.7408 | 0.8327 | 0.6824 | 0.8078 | 0.709 | 0.5593 | 0.715 | 0.5177 | 0.7431 |
| | RMSprop | 0.5778 | 0.699 | 0.5191 | 0.7437 | 0.6483 | 0.6673 | 0.5329 | 0.7404 | 0.81 | 0.6978 | 0.805 | 0.7061 | 0.6444 | 0.6686 | 0.5286 | 0.7431 |
| | SGD | 0.6268 | 0.6515 | 0.6102 | 0.6647 | 0.9838 | 0.5811 | 1.2828 | 0.515 | 0.892 | 0.6198 | 0.8765 | 0.6413 | 1.0046 | 0.584 | 0.6473 | 0.6679 |
| Inception V3 | Adam | 0.0249 | 0.9923 | 0.0699 | 0.9861 | 0.0548 | 0.9905 | 0.1228 | 0.9854 | 0.5129 | 0.9871 | 0.5111 | **0.9886** | 0.0551 | 0.9896 | 0.1076 | 0.9859 |
| | Adamax | 0.0225 | 0.9923 | 0.0375 | 0.9882 | 0.0196 | 0.9931 | 0.0494 | 0.9856 | 0.5105 | 0.9904 | 0.5124 | **0.9886** | 0.0214 | 0.9923 | 0.0447 | 0.9872 |
| | RMSprop | 0.0319 | 0.9915 | 0.0689 | 0.9856 | 0.0467 | 0.9905 | 0.1818 | 0.9784 | 0.5098 | 0.9906 | 0.513 | 0.9872 | 0.0453 | 0.9907 | 0.1115 | 0.9837 |
| | SGD | 0.0427 | 0.9851 | 0.0395 | 0.9875 | 0.0333 | 0.9876 | 0.041 | 0.987 | 0.5186 | 0.9841 | 0.5162 | 0.9864 | 0.0349 | 0.9871 | 0.0397 | 0.9874 |
| Incption ResNetV2 | Adam | 0.0263 | 0.9932 | 0.0435 | 0.9896 | 0.0382 | 0.9927 | 0.0795 | 0.9891 | 0.5101 | 0.9897 | 0.513 | 0.9874 | 0.0363 | 0.992 | 0.0723 | 0.9891 |
| | Adamax | 0.0201 | 0.9933 | 0.0341 | 0.9896 | 0.0217 | 0.9924 | 0.0349 | **0.9902** | 0.5082 | 0.9926 | 0.5104 | 0.9901 | 0.02 | 0.9931 | 0.0387 | 0.9885 |
| | RMSprop | 0.0238 | 0.9929 | 0.0493 | 0.9891 | 0.03 | 0.9932 | 0.0805 | 0.9894 | 0.5088 | 0.9912 | 0.5108 | 0.9893 | 0.0344 | 0.9929 | 0.115 | 0.9854 |
| | SGD | 0.0353 | 0.9882 | 0.0359 | 0.9891 | 0.0294 | 0.9896 | 0.0347 | 0.989 | 0.5142 | 0.9878 | 0.5136 | 0.9877 | 0.0283 | 0.9902 | 0.0368 | 0.9882 |
| MobileNet | Adam | 0.0205 | 0.9945 | 0.0779 | 0.9869 | 0.0383 | 0.9925 | 0.1537 | 0.9861 | 0.5122 | 0.9878 | 0.5128 | 0.9872 | 0.0417 | 0.9926 | 0.1132 | 0.9872 |
| | Adamax | 0.0218 | 0.9927 | 0.0394 | 0.9891 | 0.0209 | 0.9926 | 0.0525 | 0.987 | 0.5109 | 0.9899 | 0.512 | **0.9893** | 0.0177 | 0.9929 | 0.0624 | 0.985 |
| | RMSprop | 0.0178 | 0.995 | 0.0857 | 0.9845 | 0.032 | 0.9923 | 0.1128 | 0.9864 | 0.5085 | 0.9919 | 0.5122 | 0.9877 | 0.0307 | 0.9934 | 0.098 | 0.9869 |
| | SGD | 0.048 | 0.984 | 0.0457 | 0.9866 | 0.0417 | 0.9851 | 0.0477 | 0.9851 | 0.523 | 0.9792 | 0.5172 | 0.9846 | 0.0385 | 0.9858 | 0.0467 | 0.987 |
| MobileNet V2 | Adam | 0.0328 | 0.9903 | 0.0942 | 0.9816 | 0.0694 | 0.9869 | 0.1901 | 0.9766 | 0.5167 | 0.983 | 0.5168 | 0.9832 | 0.0745 | 0.9858 | 0.1636 | 0.9782 |
| | Adamax | 0.0273 | 0.9907 | 0.0524 | 0.983 | 0.0272 | 0.9902 | 0.0674 | 0.9822 | 0.5134 | 0.9877 | 0.5171 | 0.9832 | 0.0283 | 0.9894 | 0.0805 | 0.9766 |
| | RMSprop | 0.0383 | 0.9884 | 0.0878 | **0.9838** | 0.0582 | 0.9882 | 0.1599 | 0.9805 | 0.5142 | 0.9859 | 0.5163 | 0.9835 | 0.0551 | 0.9883 | 0.1773 | 0.9763 |
| | SGD | 0.0556 | 0.9799 | 0.0512 | 0.9822 | 0.051 | 0.9808 | 0.0559 | 0.9816 | 0.5254 | 0.9768 | 0.5216 | 0.9803 | 0.0478 | 0.9824 | 0.0588 | 0.9795 |
| DenseNet 121 | Adam | 0.0306 | 0.9893 | 0.0595 | 0.9867 | 0.0483 | 0.989 | 0.1444 | 0.9781 | 0.5139 | 0.9864 | 0.516 | 0.9838 | 0.0539 | 0.9886 | 0.0918 | 0.9866 |
| | Adamax | 0.0282 | 0.9894 | 0.0437 | 0.985 | 0.0267 | 0.9906 | 0.0447 | 0.9861 | 0.5126 | 0.9885 | 0.5163 | 0.9842 | 0.0263 | 0.9905 | 0.0449 | 0.984 |
| | RMSprop | 0.0333 | 0.9889 | 0.0693 | 0.9853 | 0.0422 | 0.9892 | 0.0984 | 0.9866 | 0.5133 | 0.9868 | 0.5159 | 0.9843 | 0.051 | 0.9882 | 0.0944 | 0.9853 |
| | SGD | 0.0468 | 0.9833 | 0.0398 | **0.9869** | 0.0388 | 0.9846 | 0.0433 | 0.9859 | 0.5237 | 0.9794 | 0.5163 | 0.985 | 0.0384 | 0.9858 | 0.0403 | 0.9854 |
| NASNet Large | Adam | 0.0135 | 0.9991 | 0.1179 | 0.9949 | 0.029 | 0.9985 | 0.2231 | 0.9947 | 0.5041 | 0.9959 | 0.5044 | 0.9957 | 0.0492 | 0.9965 | 0.0785 | 0.9947 |
| | Adamax | 0.0016 | 0.9996 | 0.0185 | **0.9965** | 0.0017 | 0.9993 | 0.0231 | 0.9954 | 0.5013 | 0.9988 | 0.5038 | 0.9962 | 0.0024 | 0.9993 | 0.0217 | 0.9955 |
| | RMSprop | 0.0052 | 0.9993 | 0.0674 | 0.9949 | 0.0134 | 0.9987 | 0.1032 | 0.9954 | 0.5021 | 0.998 | 0.5048 | 0.995 | 0.0136 | 0.9988 | 0.0484 | 0.9962 |
| | SGD | 0.0103 | 0.9971 | 0.0168 | 0.9957 | 0.0061 | 0.998 | 0.0159 | 0.9955 | 0.5053 | 0.9955 | 0.5052 | 0.9955 | 0.0058 | 0.9981 | 0.0184 | 0.9957 |

TABLE 7
PRECISION, RECALL, AND F1-SCORES OF THE BEST COMBINATION FOR EVERY MODELS

| Model | | Precision | Recall | f1-score | Model | | Precision | Recall | f1-score |
|---|---|---|---|---|---|---|---|---|---|
| Xception | Cat | 0.992 | 0.986 | 0.9891 | Inception ResNetV2 | Cat | 0.9932 | 0.987 | 0.9902 |
| | Dog | 0.9863 | 0.992 | 0.9892 | | Dog | 0.9873 | 0.993 | 0.9903 |
| | **Accuracy** | | | 0.9891 | | **Accuracy** | | | 0.9902 |
| | **Macro-F1** | | | 0.9891 | | **Macro-F1** | | | 0.9902 |
| | **Weighted-F1** | | | 0.9891 | | **Weighted-F1** | | | 0.9902 |
| VGG16 | Cat | 0.9504 | 0.932 | 0.9409 | MobileNet | Cat | 0.9916 | 0.987 | 0.9893 |
| | Dog | 0.9329 | 0.951 | 0.942 | | Dog | 0.9869 | 0.992 | 0.9893 |
| | **Accuracy** | | | 0.9414 | | **Accuracy** | | | 0.9893 |
| | **Macro-F1** | | | 0.9414 | | **Macro-F1** | | | 0.9893 |
| | **Weighted-F1** | | | 0.9414 | | **Weighted-F1** | | | 0.9893 |
| VGG19 | Cat | 0.9204 | 0.947 | 0.9335 | MobileNet V2 | Cat | 0.9785 | 0.989 | 0.9839 |
| | Dog | 0.9453 | 0.918 | 0.9315 | | Dog | 0.9893 | 0.978 | 0.9837 |
| | **Accuracy** | | | 0.9325 | | **Accuracy** | | | 0.9838 |
| | **Macro-F1** | | | 0.9325 | | **Macro-F1** | | | 0.9838 |
| | **Weighted-F1** | | | 0.9325 | | **Weighted-F1** | | | 0.9838 |
| ResNet50 | Cat | 0.7358 | 0.836 | 0.7827 | DenseNet 121 | Cat | 0.9942 | 0.98 | 0.9868 |
| | Dog | 0.8068 | 0.696 | 0.747 | | Dog | 0.9798 | 0.994 | 0.987 |
| | **Accuracy** | | | 0.7662 | | **Accuracy** | | | 0.9869 |
| | **Macro-F1** | | | 0.7648 | | **Macro-F1** | | | 0.9869 |
| | **Weighted-F1** | | | 0.765 | | **Weighted-F1** | | | 0.9869 |
| InceptionV3 | Cat | 0.9897 | 0.988 | 0.9886 | NASNet Large | Cat | 0.9974 | 0.9955 | 0.9965 |
| | Dog | 0.9875 | 0.99 | 0.9887 | | Dog | 0.9955 | 0.9974 | 0.9965 |
| | **Accuracy** | | | 0.9886 | | **Accuracy** | | | 0.9965 |
| | **Macro-F1** | | | 0.9886 | | **Macro-F1** | | | 0.9965 |
| | **Weighted-F1** | | | 0.9886 | | **Weighted-F1** | | | 0.9965 |

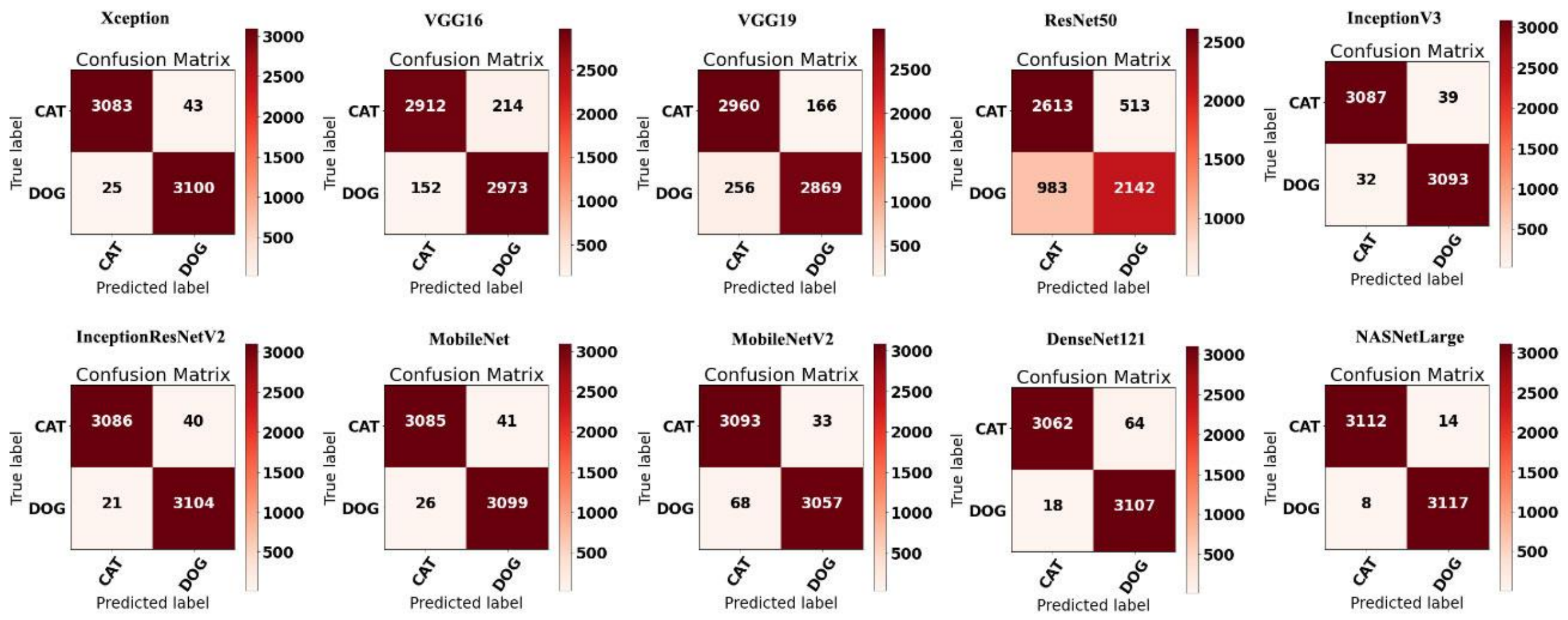

**FIGURE 8. Confusion Matrices**

TABLE 8
TRAIN & TEST TIME COMPARISON ON DIFFERENT GPU ARCHITECTURES

| GPU | NVIDIA GeForce | | | | | |
|---|---|---|---|---|---|---|
| | GTX 1070 Laptop (8GB) | | RTX 3080Ti Laptop (16GB) | | RTX 3090 Desktop (24GB) | |
| Model | Train (Sec) | Test (Sec) | Train (Sec) | Test (Sec) | Train (Sec) | Test (Sec) |
| Xception | 6210 | 50.5 | 5040 | 39.3 | 3210 | 19.8 |
| VGG16 | 6240 | 57.6 | 5400 | 40.7 | 3201 | 18.9 |
| VGG19 | 6247 | 67.1 | 5640 | 46.2 | 3205 | 19.4 |
| ResNet50 | 6002 | 47.5 | 5040 | 39.1 | 3300 | 19.7 |
| InceptionV3 | 5850 | 40.3 | 4200 | 44.9 | 3352 | 19.9 |
| InceptionResNetV2 | 6900 | 70.2 | 4800 | 45.5 | 3360 | 21.9 |
| MobileNet | 5867 | 39.1 | 5592 | 43.2 | 3180 | 19.2 |
| MobileNetV2 | 5899 | 39.1 | 5595 | 44.8 | 3240 | 19.4 |
| DenseNet121 | 5912 | 42.9 | 5901 | 44.7 | 3356 | 20.4 |
| NASNetLarge | 18540 | 378 | 13800 | 210 | 7320 | 96.8 |

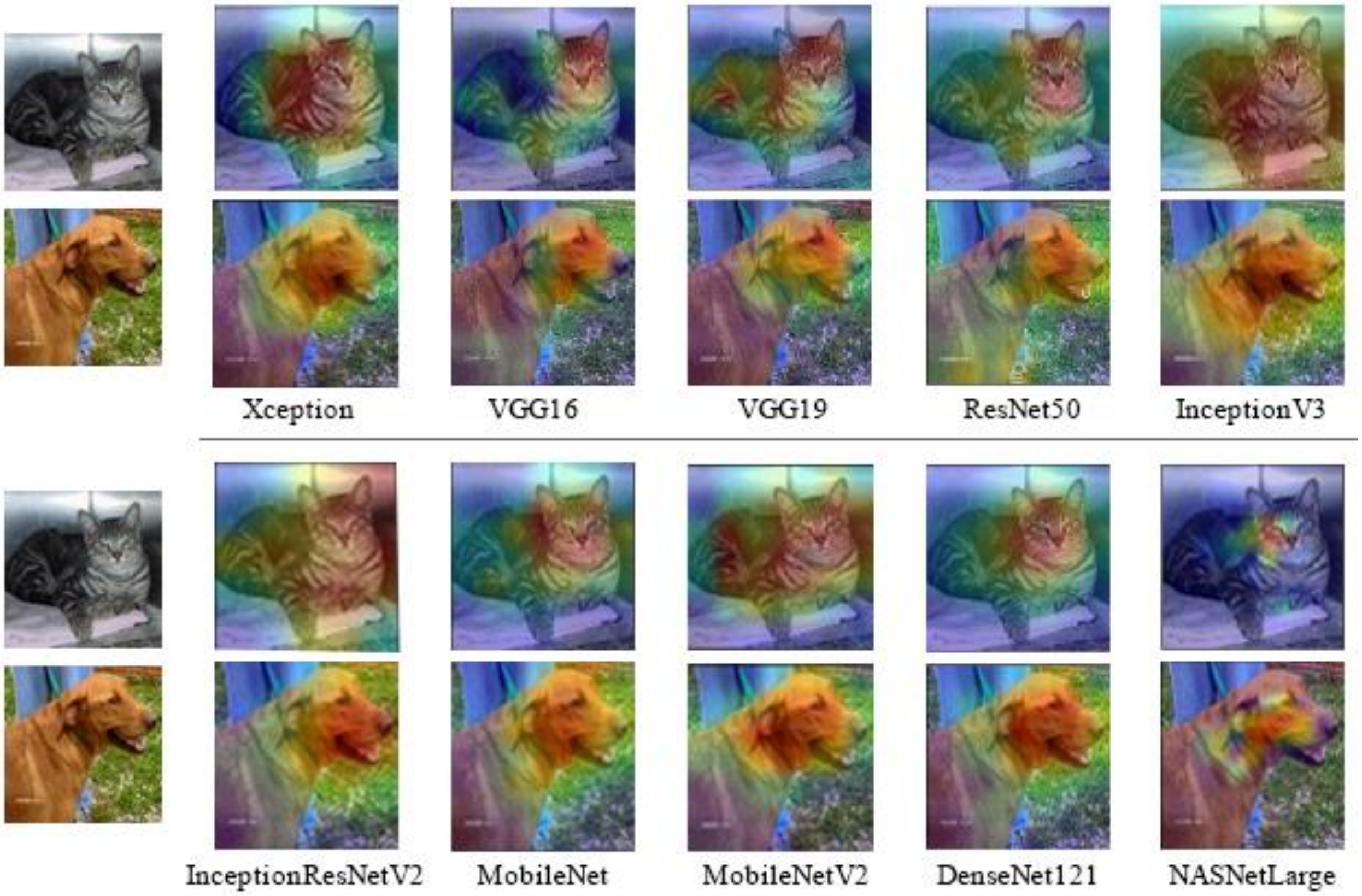

FIGURE 9. **Gradient-weighted Class Activation Mapping for all models**

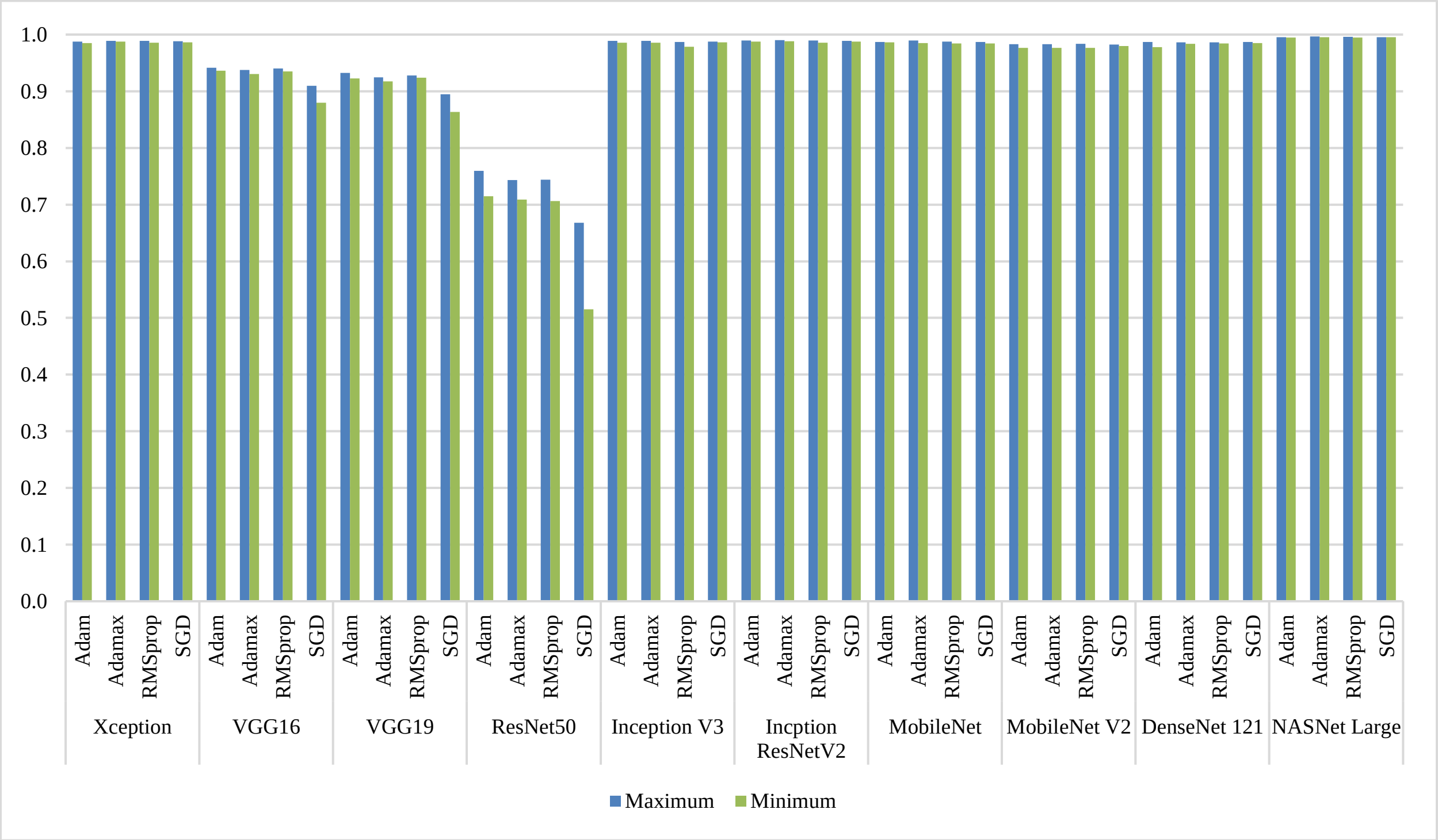

FIGURE 10. **Optimizer Comparison**

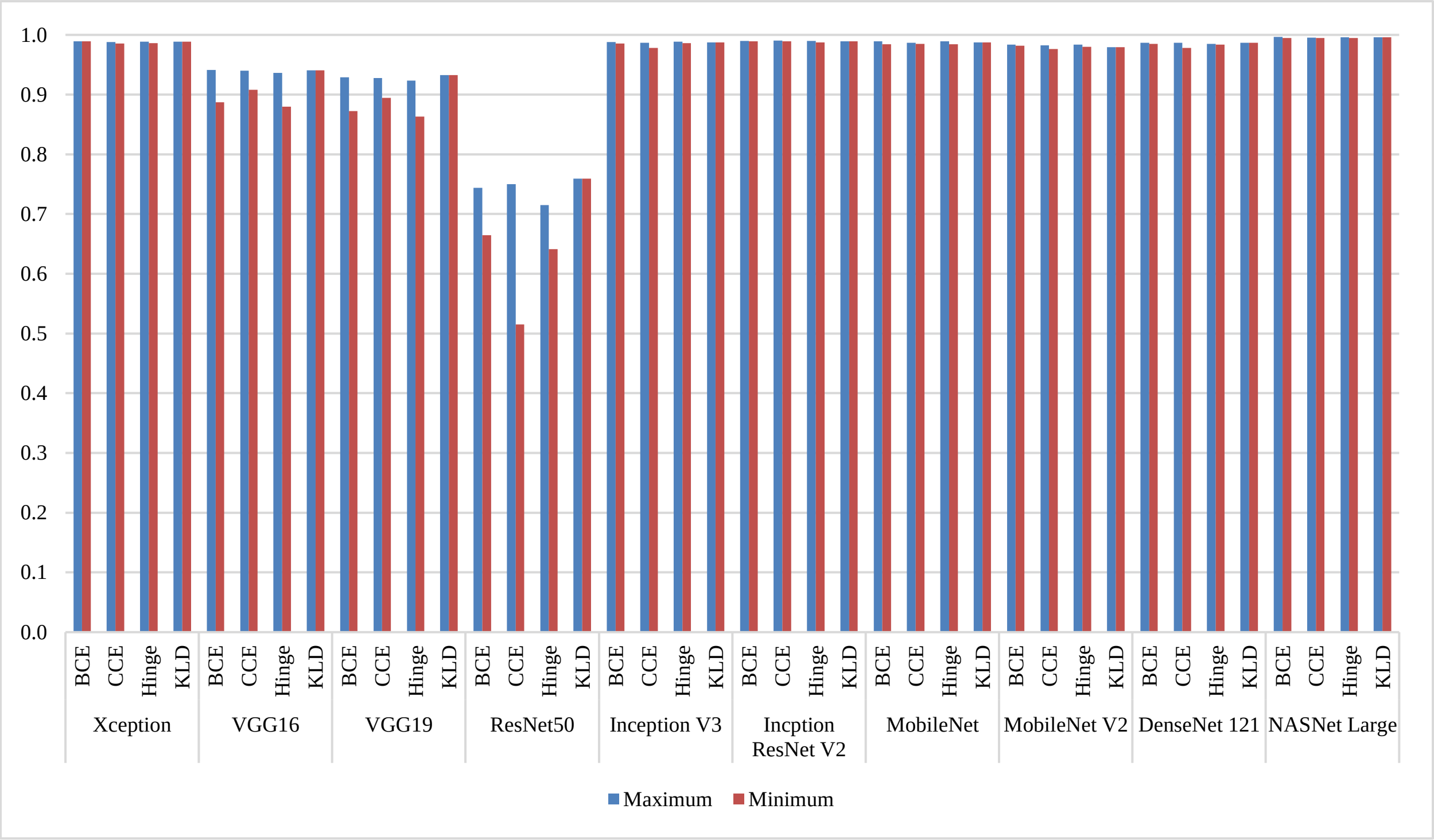

**FIGURE 11.** Loss Function Comparison

### *I. OPTIMIZER & LOSS FUNCTION COMPARISON*

In **Figure 10**, a comparison has been shown for the highest and lowest accuracy range for all the optimizers used in the models. It should be noted that in most cases, the Adamax optimizer provides the highest accuracy and SGD provides the lowest. In **Figure 11**, a comparison has been shown for the highest and lowest accuracy range for all the loss functions used in the models. It is worth mentioning that in most cases, the BCE provides the highest accuracy.

### *J. GPU PROCESSING TIME COMPARISON*

In terms of processing time for all GPU architectures, there is no specific pattern followed. The un-uniform pattern is mainly due to the inherent structure of GPUs. **Table 8** shows train and test comparison on different GPU architectures.

## V. CONCLUSION

The development of more complex Deep Neural Networks (DNNs) has become central in advancing the state-of-the-art for ImageNet-1k competition and other vision tasks. In this article, we compare the performance of various pre-trained models to provide a useful tool for selecting an appropriate architecture based on resource constraints in practical applications. We examine the top 10 pre-trained models on the ASIRRA Cats and Dogs dataset in terms of accuracy, the number of parameters, different optimizers, and loss functions, as well as GPU processing time. We can verify from this experiment that using the NASNetLarge model with Adamax Optimizer and Binary Cross Entropy loss function achieves the highest accuracy of about 99.65% which has outperformed all previous results from other experiments on this dataset. As a result, we can say that our obtained results can be considered a state-of-the-art achievement.

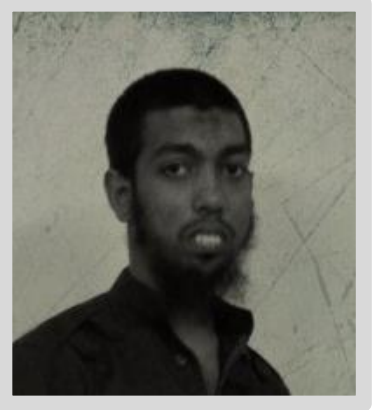

**GALIB MUHAMMAD SHAHRIAR HIMEL** received his 1st BSc degree in Computer Science & Engineering from Ahsanullah University of Science and Technology (AUST) in the year 2016. Then he received his 1st MSc degree in Computer Science & Engineering from United International University (UIU) in the year 2018. Then he received his 2nd BSc degree in Computing from the University of Greenwich (UoG), UK in 2021. After that, he received his 2nd MSc degree in Computer Science specializing in Intelligent Systems from American International University-Bangladesh (AIUB) in the year 2022. He has completed his 3rd MSc degree in Applied Physics and Electronics from Jahangirnagar University (JU) in the year 2023. He has worked as a researcher at the Bangladesh University of Business and Technology (BUBT) and also a part-time researcher at Independent University, Bangladesh (IUB). He is also involved in several types of research related to Bio-medical image processing using machine learning. His research interest includes Artificial Intelligence, Machine Learning, Bioinformatics, Bio-medical image analysis & Computer Vision. Currently, He is pursuing his PhD degree at Universiti Sains Malaysia.

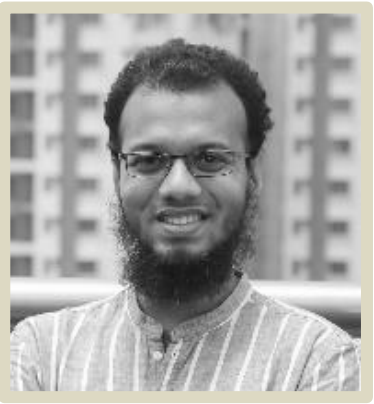

**MD. MASUDUL ISLAM,** an Academician, Researcher & former In-House Web Developer from Bangladesh. He has been working as a teacher at Bangladesh University of Business & Technology (BUBT) in the department of CSE from 2013 till now. At present, he is continuing as Assistant Professor, Dept. of CSE in BUBT. He is doing Ph.D. in the Department of CSE, at Jahangirnagar University. He loves everything that has to do with Teaching, IT-Software, Data Science, Web Programming, Astronomy, Database, Quantum Computing, History, Religion, and System Analysis. He has a true devotion to teaching and research.